\documentclass[11pt,a4paper]{article}
\usepackage[hyperref]{acl2018}
\usepackage{times}
\usepackage{latexsym}

\usepackage{url}

\aclfinalcopy 
\def\aclpaperid{881} 

\usepackage{amsmath}
\usepackage{tabularx}
\usepackage{booktabs}
\usepackage{amssymb}
\usepackage{pifont}
\usepackage{bm}
\usepackage{bbm}
\usepackage{multirow}
\usepackage{tcolorbox}
\usepackage[framemethod=tikz]{mdframed}
\usepackage{tikz,pgfplots}
\usepackage{xspace}
\usepgfplotslibrary{groupplots}
\pgfplotsset{compat=1.13}

\newcommand{\V}[1]{\bm{#1}}

\definecolor{g-red}{HTML}{DB4437}
\definecolor{g-blue}{HTML}{4285F4}
\definecolor{g-green}{HTML}{0F9D58}
\definecolor{g-yellow}{HTML}{F4B400}
\definecolor{g-orange}{HTML}{FF9800}
\definecolor{g-grey}{HTML}{9E9E9E}
\definecolor{uw}{RGB}{138,43,226}
\definecolor{stanford}{RGB}{255,69,0}
\definecolor{const}{RGB}{68, 110, 182}
\definecolor{head}{RGB}{246, 180, 32}
\definecolor{freq}{RGB}{0, 0, 0}

\newmdenv[innerlinewidth=0.5pt, roundcorner=4pt,linecolor=black,innerleftmargin=6pt,
innerrightmargin=6pt,innertopmargin=6pt,innerbottommargin=6pt]{examplebox}

\newcommand\layerbox[4]{
\draw[rounded corners] (#2, #3) rectangle (#2 + #1 * #4, #3 + #1 * 1);
}

\newcommand\layercolorbox[5][0.4] {
\draw[rounded corners, fill=#5] (#2, #3) rectangle (#2 + #1 * #4, #3 + #1 * 1);
}

\newcommand\layercomponent[5]{
\filldraw[fill=#5] (#2 + #1 * #4 - #1 * 0.5, #3 + #1 * 0.5) circle (#1 * 0.4);
}
\newcommand\layer[5][0.4] {
\layerbox{#1}{#2}{#3}{#4}
\foreach \x in {1, ..., #4}{
  \layercomponent{#1}{#2}{#3}{\x}{#5}
}
}

\newcommand\sumnode[3] {
\layercolorbox{#1-0.2}{#2-0.2}{1}{#3}
\node at (#1, #2) {\textcolor{white}{+}};
}

\newcommand\archcomment[3] {
\node[anchor=west, align=left] at (-5.5, #1) {\textbf{#2}};
}

\newcommand\archcommenttwo[4] {
\archcomment{#1}{#2}{#4}
\archcomment{#1-0.4}{#3}{#4}
}

\title{Jointly Predicting Predicates and Arguments\\ in Neural Semantic Role Labeling}

\author{Luheng He \qquad Kenton Lee \qquad Omer Levy \qquad Luke Zettlemoyer\\
  Paul G. Allen School of Computer Science \& Engineering \\
  University of Washington, Seattle WA \\
  {\tt \{luheng, kentonl, omerlevy, lsz\}@cs.washington.edu}}
\date{}

\date{}

\begin{document}

\maketitle
\begin{abstract}
Recent BIO-tagging-based neural semantic role labeling models are 
very high performing, but assume gold predicates as part of the input and cannot incorporate span-level features. 
We propose an end-to-end approach for jointly predicting all predicates, arguments spans, and the relations between them. 
The model makes independent decisions about what relationship, if any, holds between every possible word-span pair, and learns contextualized span representations that provide rich, shared input features for each decision. 
Experiments demonstrate that this approach sets a new state of the art on PropBank SRL without gold predicates.\footnote{Code and models: \href{https://github.com/luheng/lsgn}{https://github.com/luheng/lsgn}}
\end{abstract}

\section{Introduction}

\begin{figure}[t]
\centering
\includegraphics[width=0.87\columnwidth, keepaspectratio,trim={0.5cm 2.2cm 0.4cm 0cm},clip]{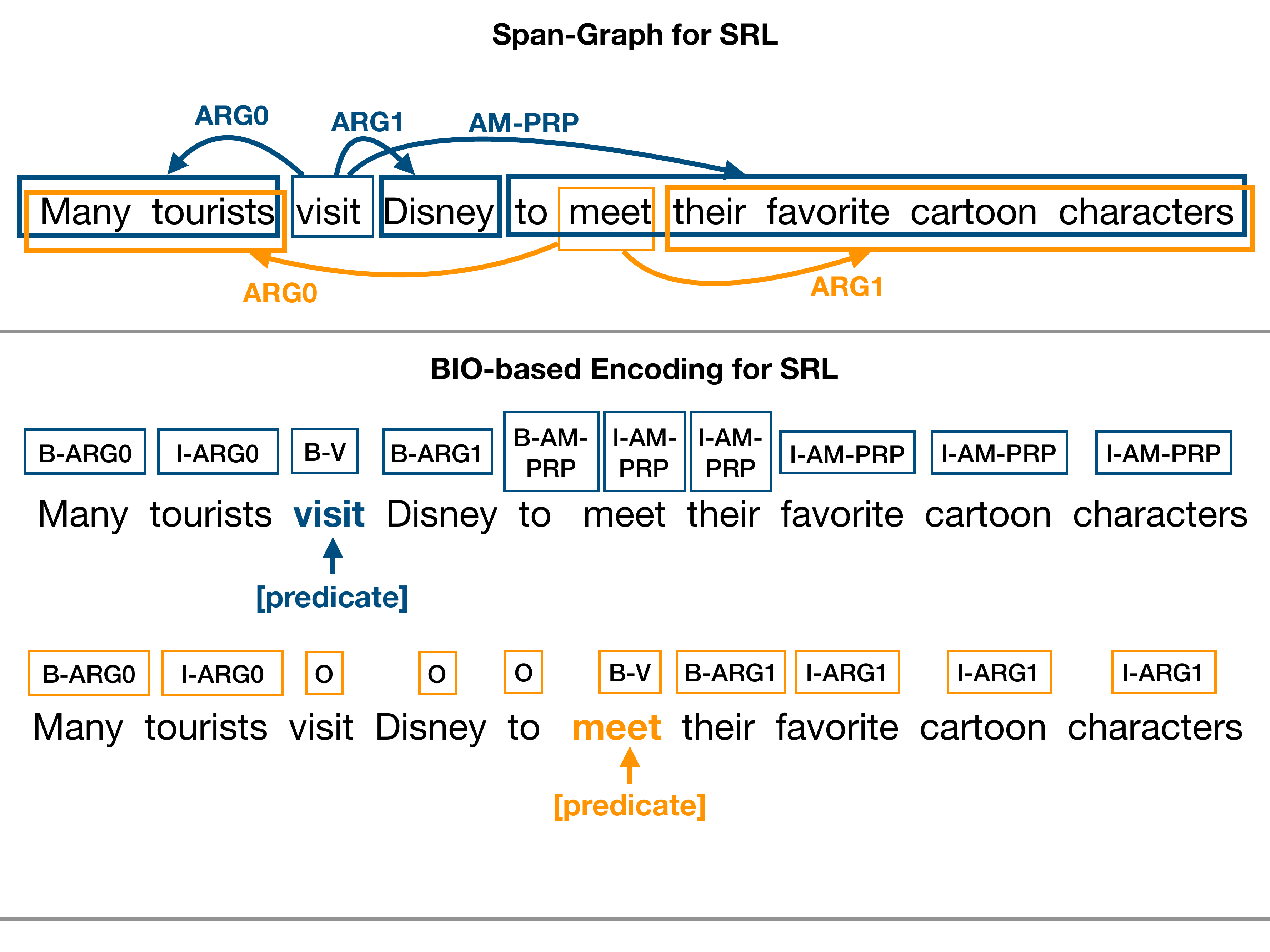}
\vspace{-1.5em}
\caption{
A comparison of our span-graph structure (top) versus BIO-based SRL (bottom).}\label{fig:intro_example}
\vspace{-1em}
\end{figure}

Semantic role labeling (SRL) captures predicate-argument relations, such as ``who did what to whom.''
Recent high-performing SRL models \cite{He2017DeepSR,Marcheggiani2017ASA, tan2018deep} are BIO-taggers, labeling argument spans for a single predicate at a time (as shown in Figure \ref{fig:intro_example}).
They are typically only evaluated with gold predicates, and must be pipelined with error-prone predicate identification models for deployment.

We propose an end-to-end approach for predicting all the predicates and their argument spans in one forward pass.
Our model builds on a recent coreference resolution model \cite{Lee2017EndtoendNC}, by making central use of learned, contextualized span representations. We use these representations to predict SRL graphs directly over text spans. 
Each edge is identified by independently predicting which role, if any, holds between every possible pair of text spans, while using aggressive beam pruning for efficiency. The final graph is simply the union of predicted SRL roles (edges) and their associated text spans (nodes). 

Our span-graph formulation overcomes a key limitation of semi-markov and BIO-based models~\cite{kong2015segmental,zhou2015end,Yang2017AJS,He2017DeepSR,tan2018deep}: it can model overlapping spans across different predicates in the same output structure (see Figure \ref{fig:intro_example}).
The span representations also generalize the token-level representations in BIO-based models, letting the model dynamically decide which spans and roles to include, without using previously standard syntactic features~\cite{punyakanok2008importance,fitzgerald2015semantic}. 

To the best of our knowledge, this is the first span-based SRL model that does not assume that predicates are given. In this more realistic setting, where the predicate must be predicted, our model achieves state-of-the-art performance on PropBank.
It also reinforces the strong performance of similar span embedding methods for coreference~\cite{Lee2017EndtoendNC}, suggesting that this style of models could be used for other span-span relation tasks, 
such as syntactic parsing~\cite{stern2017minimal},
relation extraction~\cite{miwa2016end}, and QA-SRL~\cite{fitzgerald2018large}.

\section{Model}\label{sec:model}

We consider the space of possible predicates to be all the tokens in the input sentence, and the space of arguments to be all continuous spans.
Our model decides what relation exists between each predicate-argument pair (including \textit{no relation}).

Formally, given a sequence $X=w_1,\dots,w_n$, we wish to predict a set of labeled predicate-argument relations $Y \subseteq \mathcal{P}\times \mathcal{A}\times\mathcal{L}$,
where $\mathcal{P}=\{w_1,\ldots, w_n\}$ is the set of all tokens (predicates), $\mathcal{A} = \{(w_i,\dots,w_j) \mid 1 \leq i \leq j \leq n \}$ contains all the spans (arguments), and $\mathcal{L}$ is the space of semantic role labels, including a null label $\epsilon$ indicating no relation. The final SRL output would be all the non-empty relations $\{(p,a,l)\in Y \mid l \neq \epsilon\}$.

We then define a set of random variables, where each random variable $y_{p,a}$ corresponds to a predicate $p\in\mathcal{P}$ and an argument $a\in\mathcal{A}$, 
taking value from the discrete label space $\mathcal{L}$. The random variables $y_{p,a}$ are conditionally independent of each other given the input $X$:
\begin{align}
    P(Y\mid X) =& \prod_{p\in\mathcal{P},a\in\mathcal{A}} P(y_{p,a} \mid X) \\
    P(y_{p,a}=l\mid X) =& \frac{\exp(\phi(p, a, l))}{\sum\limits_{l'\in \mathcal{L}}\exp(\phi(p, a, l'))}
\end{align}
Where $\phi(p,a,l)$ is a scoring function for a possible (predicate, argument, label) combination.
$\phi$ is decomposed into two unary scores on the predicate and the argument (defined in  Section \ref{sec:arch}), as well as a label-specific score for the relation:
\begin{align}
    \phi(p, a, l) 
    =  \Phi_{\text{a}}(a) + \Phi_{\text{p}}(p) + \Phi_{\text{rel}}^{(l)}(a, p) \label{eq:phi:tenary}
\end{align}
The score for the \textit{null} label is set to a constant:
$\phi(p, a, \epsilon) = 0$, similar to  logistic regression.

\paragraph{Learning}
For each input $X$, we minimize the negative log likelihood of the gold structure $Y^*$:
\begin{align}
    \mathcal{J}(X) =& - \log P (Y^* \mid X) \label{eq:loss}
\end{align}

\paragraph{Beam pruning}
As our model deals with $O(n^2)$ possible argument spans and $O(n)$ possible predicates, it needs to consider $O(n^3|\mathcal{L}|)$ possible relations, which is computationally impractical. To overcome this issue, we define two beams $B_{\text{a}}$ and $B_{\text{p}}$ for storing the candidate arguments and predicates, respectively. The candidates in each beam are ranked by their unary score ($\Phi_{\text{a}}$ or $\Phi_{\text{p}}$). The sizes of the beams are limited by $\lambda_{\text{a}}n$ and $\lambda_{\text{p}}n$. Elements that fall out of the beam do not participate in computing the edge factors $\Phi_{\text{rel}}^{(l)}$, reducing the overall number of relational factors evaluated by the model to  $O(n^2|\mathcal{L}|)$.
We also limit the maximum width of spans to a fixed number $W$ (e.g. $W=30$), further reducing the number of computed unary factors to $O(n)$.

\newcommand\lstmnode[2] {
\filldraw[fill=g-yellow] (#1+0.2,#2+0.3) rectangle (#1+0.6,#2+0.7); 
}

\newcommand\forwardlstmconnect[3] {
\draw[-latex, line width=1pt] (#1+0.6,#2+0.25+0.7) -- (#1-0.3+#3,#2+0.25+0.7);
}
\newcommand\backwardlstmconnect[3] {
\draw[-latex, line width=1pt] (#1-0.3+#3,#2+0.25+0.7) -- (#1+0.6,#2+0.25+0.7);
}
\newcommand\lstmconnect[3] {
\forwardlstmconnect{#1}{#2+0.35}{#3}
\backwardlstmconnect{#1}{#2 + 1.15}{#3}
}

\newcommand\spanout[4] {
\node[anchor=north, align=center] at (#1+0.6, 5.6) {$\text{\small #4}$}; 
\layer{#1}{4.6}{3}{g-green} 
\draw[-latex, line width=1pt, out=90, in=-90] (1.5 * #2 +0.4, 3.1) to (#1+0.2, 4.6); 
\draw[-latex, line width=1pt, out=90, in=-90] (1.5 * #3 + 0.4, 3.1) to (#1+1, 4.6); 
\sumnode{#1 + 0.6}{4.1}{g-red}
\foreach \x in {#2, ..., #3}{
\draw[-latex, line width=1pt, in=-90, out=90, looseness=1] (1.5 * \x + 0.4, 3.1) to (#1+0.6, 3.9); 
}
\draw[-latex, line width=1pt, in=-90, out=90, looseness=1] (#1+0.6, 4.3) to (#1+0.6, 4.6); 
}

\newcommand\mentionout[4] {
\node[anchor=north, align=center] at (#1+0.6, 7.2) {$\text{\small #4}$}; 
\layer{#1}{4.6}{3}{g-green} 
\layer{#1+0.4}{6.3}{1}{black} 
\draw[-latex, line width=1pt] (#1+0.6, 5) to (#1+0.6, 6.3); 

\draw[-latex, line width=1pt, out=90, in=-90] (1.5 * #2 +0.4, 3.1) to (#1+0.2, 4.6); 

\draw[-latex, line width=1pt, out=90, in=-90] (1.5 * #3 + 0.4, 3.1) to (#1+1, 4.6); 

\sumnode{#1 + 0.6}{4.1}{g-red}

\foreach \x in {#2, ..., #3}{
\draw[-latex, line width=1pt, in=-90, out=90, looseness=1] (1.5 * \x + 0.4, 3.1) to (#1+0.6, 3.9); 
}
\draw[-latex, line width=1pt, in=-90, out=90] (#1+0.6, 4.3) to (#1+0.6, 4.6); 
}

\newcommand\predicateout[3] {
\node[anchor=north, align=center] at (#1+0.4, 7.2) {$\textit{\textbf{\small #3}}$}; 
\layer{#1}{4.6}{2}{cyan} 
\layer{#1+0.2}{6.3}{1}{g-grey} 
\draw[-latex, line width=1pt] (#1+0.4, 5) to (#1+0.4, 6.3); 

\draw[-latex, line width=1pt, out=90, in=-90] (1.5 * #2 +0.4, 3.1) to (#1+0.4, 4.6); 
}

\newcommand\srlout[4] {
    \layer{#1-0.4}{5.6}{3}{white} 
    \node[anchor=north, align=center] at (#1+0.2, 6.4) {$\textbf{\small #4}$}; 
    
    \draw[-latex, line width=1pt, out=90, in=-90] (#2 +0.6, 5) to (#1+0.2, 5.6); 
    \draw[-latex, line width=1pt, out=90, in=-90] (#3 +0.4, 5) to (#1+0.2, 5.6); 
}

\newcommand\word[2] {
\node[anchor=mid] at (#1+0.4, 0) {\small #2}; 
\layer{#1}{0.3}{2}{g-blue} 

\lstmnode{#1}{0.8}
\lstmnode{#1}{1.6}
\draw[-latex, line width=1pt] (#1+0.4, 0.7) to (#1+0.4, 1.1); 
\draw[-latex, line width=1pt, out=150, in=-150, looseness=1.5] (#1+0.4,0.7) to (#1+0.4, 1.9); 

\draw[-latex, line width=1pt, out=30, in=-30, looseness=1.5] (#1+0.4, 2 - 0.5) to (#1+0.4, 2 + 0.7); 

\draw[-latex, line width=1pt] (#1+0.4, 2.3) to (#1+0.4, 2.7); 

\layer{#1}{2.7}{2}{g-yellow} 
}

\begin{figure*}[ht!]
\begin{centering}
\scalebox{0.75} {
\begin{tikzpicture}
\word{-1.5}{Many}
\lstmconnect{-1.5}{0}{2}
\word{0}{tourists}
\lstmconnect{0}{0}{2}
\word{1.5}{visit}
\lstmconnect{1.5}{0}{2}
\word{3}{Disney}
\lstmconnect{3}{0}{2}
\word{4.5}{to}
\lstmconnect{4.5}{0}{2}
\word{6}{meet}
\lstmconnect{6}{0}{2}
\word{7.5}{their}
\lstmconnect{7.5}{0}{2}
\word{9}{favorite}
\lstmconnect{9}{0}{2}
\word{10.5}{cartoon}
\lstmconnect{10.5}{0}{2}
\word{12}{characters}

\spanout{-1.0}{-1}{0}{Many tourists}
\spanout{1.3}{0}{2}{tourists visit Disney}
\spanout{3.5}{2}{3}{Disney to}
\spanout{5.0}{3}{4}{to meet}
\spanout{7.3}{4}{6}{meet their favorite}
\spanout{10.3}{6}{8}{favorite cartoon characters}


\archcomment{4.8}{Span representation ($\V{g}$)}{g-green}
\archcomment{4.1}{Span head ($\mathbf{x}_{\text{h}}$)}{g-red}
\archcommenttwo{2.9}{Bidirectional LSTM}{($\mathbf{\bar{x}}^m$)}{g-yellow}
\archcommenttwo{0.6}{Word \& character}{representation ($\mathbf{x}$)}{g-blue}
\end{tikzpicture}
}
\caption{Building the argument span representations $\mathbf{g}(a)$ from BiLSTM outputs. For clarity, we only show one BiLSTM layer and a small subset of the arguments.
}
\label{fig:model_viz}
\end{centering}
\end{figure*}
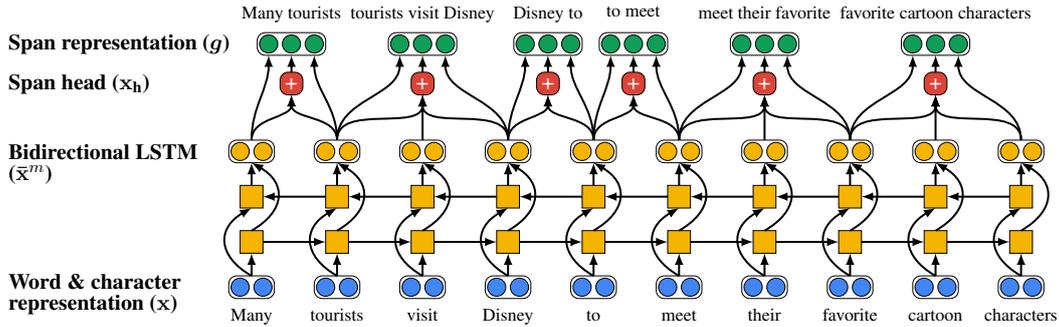

\newcommand\score[7] {
\node[anchor=#6, align=#7] at (#3, 7.9) {\small #4}; 
\node[anchor=#6, align=#7] at (#3, 7.6) {\small #5}; 
\layer{#3}{7.3}{1}{g-grey} 
\draw[-latex, line width=1pt] (#1+0.6, 5.7) to (#3+0.2, 7.3); 
\draw[-latex, line width=1pt] (#2+0.6, 5.7) to (#3+0.2, 7.3); 
\draw[-latex, line width=1pt] (#3+0.2, 6.4) to (#3+0.2, 7.3); 

\layer{#3}{6}{1}{white} 

\draw[-latex, line width=1pt] (#1+0.6, 4.9) to (#3+0.2, 6); 
\draw[-latex, line width=1pt] (#2+0.6, 4.9) to (#3+0.2, 6); 
}

\newcommand\unaryscore[7] {
\node[anchor=#6, align=#7] at (#3, 7.2) {\small #4}; 
\node[anchor=#6, align=#7] at (#3, 6.9) {\small #5}; 
\layer{#3}{7.3}{1}{g-grey} 
\draw[-latex, line width=1pt] (#3+0.2, 6.4) to (#3+0.2, 7.3); 

\layer{#3}{6}{1}{white} 

\draw[-latex, line width=1pt] (#1+0.6, 4.9) to (#3+0.2, 6); 
}

\newcommand\mentionin[4] {
\node[anchor=north, align=center] at (#1+0.7, 4.4) {\small #4}; 
\layer{#1}{4.5}{3}{g-green} 
\layer{#1+0.4}{5.3}{1}{black} 
\draw[-latex, line width=1pt] (#1+0.6, 4.9) to (#1+0.6, 5.3); 
}

\newcommand\predin[4] {
\node[anchor=north, align=center] at (#1+0.7, 4.4) {\small #4}; 
\layer{#1+0.2}{4.5}{2}{g-yellow} 
\layer{#1+0.4}{5.3}{1}{black} 
\draw[-latex, line width=1pt] (#1+0.6, 4.9) to (#1+0.6, 5.3); 
}

\newcommand\nerin[4] {
\node[anchor=north, align=center] at (#1+0.7, 4.4) {\small #4}; 
\layer{#1}{4.5}{3}{g-green} 
}

\begin{figure}[ht!]
\centering
\scalebox{0.74} {
\begin{tikzpicture}
\mentionin{-1}{0}{1}{Many tourists}
\predin{2.3}{7}{8}{meet}
\score{-1}{2.3}{0.5}{$\phi(\text{Many tourists},$}{$\text{meet},\textbf{ARG0})$}{east}{right}
\score{-1}{2.3}{1.8}{$\;\phi(\text{Many tourists},$}{$\;\;\;\;\;\;\text{meet},\textbf{ARG1})$}{west}{left}
\node[anchor=north, align=center] at (-0.4, 8.8) {\small$\phi(\text{Many tourists}, \text{meet}, \epsilon)=0$}; 
\layer{-0.3}{8.8}{1}{g-grey} 

\layer{0.9}{9}{3}{g-orange} 
\draw[-latex, line width=1pt] (0.1, 9) to (0.9,9.2);
\draw[-latex, line width=1pt] (2.0, 7.7) to (1.9,9);
\draw[-latex, line width=1pt] (0.7, 7.7) to (1.5,9);

\archcommenttwo{9.1}{Softmax}{$P(y_{p,a}=l\mid X)$}{g-orange}
\archcommenttwo{7.6}{Combined}{score $\phi(p,a,l)$}{g-grey}
\archcomment{6.2}{Label score $\Phi^{(l)}_{\text{rel}}$}{black}
\archcomment{5.5}{Unary scores $\Phi_{\text{a}},\Phi_{\text{p}}$}{black}
\archcommenttwo{4.8}{Span}{representation ($\V{g}$)}{g-green}
\end{tikzpicture}
\hspace{-0.4in}
}
\vspace{-1em}
\caption{The span-pair classifier takes in predicate and argument representations as inputs, and computes a softmax over the label space $\mathcal{L}$.
} 
\label{fig:tasks_viz}
\end{figure}
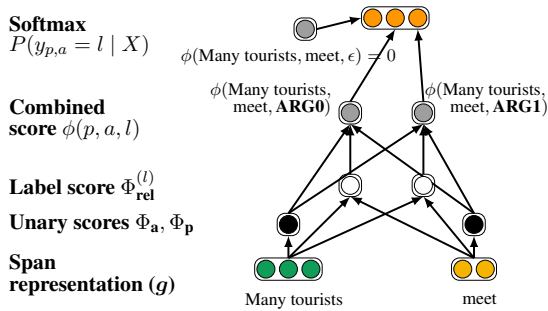

\section{Neural Architecture}\label{sec:arch}
Our model builds contextualized representations for argument spans $a$ and predicate words $p$ based on BiLSTM outputs (Figure \ref{fig:model_viz}) and uses feed-forward networks to compute the factor scores in $\phi(p,a,l)$ described in Section \ref{sec:model} (Figure \ref{fig:tasks_viz}). 

\paragraph{Word-level contexts}
The bottom layer consists of pre-trained word embeddings concatenated with character-based representations, i.e. for each token $w_i$, we have
$\mathbf{x}_i = [\textsc{WordEmb}(w_i); \textsc{CharCNN}(w_i)]$.
We then contextualize each $\mathbf{x}_i$ using an $m$-layered bidirectional LSTM with highway connections \cite{zhang2016highway}, which we denote as $\mathbf{\bar{x}}_i$.

\paragraph{Argument and predicate representation}
We build contextualized representations for all candidate arguments $a\in\mathcal{A}$ and predicates $p\in\mathcal{P}$.
The argument representation contains the following: end points from the BiLSTM outputs ($\mathbf{\bar{x}}_{\textsc{Start}(a)}, \mathbf{\bar{x}}_{\textsc{End}(a)}$), a soft head word $\mathbf{x}_{\text{h}}(a)$, and embedded span width features $\mathbf{f}(a)$, similar to \newcite{Lee2017EndtoendNC}. The predicate representation is simply the BiLSTM output at the position $\textsc{Index}(p)$.
\begin{align}
    \mathbf{g}(a) =& [\mathbf{\bar{x}}_{\textsc{Start}(a)}; \mathbf{\bar{x}}_{\textsc{End}(a)};
    \mathbf{x}_{\text{h}}(a); \mathbf{f}(a)] \\
    \mathbf{g}(p) =& \mathbf{\bar{x}}_{\textsc{Index}(p)}
\end{align}
The soft head representation $\mathbf{x}_{\text{h}}(a)$ is an attention mechanism over word inputs $\mathbf{x}$ in the argument span, where the weights $\mathbf{e}(a)$ are computed via a linear layer over the BiLSTM outputs $\mathbf{\bar{x}}$.
\begin{align}
    &\mathbf{x}_{\text{h}}(a) =  \mathbf{x}_{\textsc{Start}(a):\textsc{End}(a)}\mathbf{e}(s)^\intercal \\
    &\mathbf{e}(a) = \textsc{SoftMax} (\mathbf{w}_{\text{e}}^\intercal  \mathbf{\bar{x}}_{\textsc{Start}(a):\textsc{End}(a)})
\end{align}
$\mathbf{x}_{\textsc{Start}(a):\textsc{End}(a)}$ is a shorthand for stacking a list of vectors $\mathbf{x}_t$, where $\textsc{Start}(a)\leq t \leq \textsc{End}(a)$.

\paragraph{Scoring}
The scoring functions $\Phi$ are implemented with feed-forward networks based on the predicate and argument representations $\mathbf{g}$:
\begin{align}
     \Phi_{\text{a}} (a) =& \mathbf{w}_{\text{a}}^\intercal \text{MLP}_{\text{a}} (\mathbf{g}(a)) \\
     \Phi_{\text{p}} (p) =& \mathbf{w}_{\text{p}}^\intercal \text{MLP}_{\text{p}} (\mathbf{g}(p)) \\
     \Phi^{(l)}_{\text{rel}} (a, p) =& \mathbf{w}^{(l)\intercal}_{\text{r}} \text{MLP}_{\text{r}} ([\mathbf{g}(a); \mathbf{g}(p)]) 
\end{align}

\section{Experiments}

We experiment on the CoNLL~2005 \cite{carreras2005introduction} and CoNLL~2012 (OntoNotes~5.0, \cite{Pradhan2013TowardsRL}) benchmarks,
using two SRL setups: \textit{end-to-end} and \textit{gold predicates}. In the \textit{end-to-end} setup, a system takes a tokenized sentence as input, and predicts all the predicates and their arguments. Systems are evaluated on the micro-averaged F1 for correctly predicting (predicate, argument span, label) tuples. For comparison with previous systems, we also report results with \textit{gold predicates}, in which the complete set of predicates in the input sentence is given as well.
Other experimental setups and hyperparameteres are listed in Appendix A.1.

\paragraph{ELMo embeddings} 
To further improve performance, we also add ELMo word representations \cite{peters2018deep} to the BiLSTM input (in the +\texttt{ELMo} rows). Since the contextualized representations ELMo provides can be applied to most previous neural systems, the improvement is orthogonal to our contribution. In Table \ref{tab:srl_results_e2e} and \ref{tab:srl_results_gold}, we organize all the results into two categories: the comparable single model systems, and the models augmented with ELMo or ensembling (in the $\texttt{PoE}$ rows).

\paragraph{End-to-end results}
As shown in Table \ref{tab:srl_results_e2e},\footnote{For the end-to-end setting on CoNLL~2012, we used a subset of the train/dev data from previous work due to noise in the dataset; the dev result is not directly comparable. See Appendix A.2 for detailed explanation.} our joint model outperforms the previous best pipeline system \cite{He2017DeepSR} by an F1 difference of anywhere between 1.3 and 6.0 in every setting. 
The improvement is larger on the Brown test set, which is out-of-domain, and the CoNLL~2012 test set, which contains nominal predicates.
On all datasets, our model is able to predict over 40\% of the sentences completely correctly.

\paragraph{Results with gold predicates}
To compare with additional previous systems, we also conduct experiments with gold predicates by constraining our predicate beam to be gold predicates only.
As shown in Table \ref{tab:srl_results_gold}, our model significantly out-performs \newcite{He2017DeepSR}, but falls short of \newcite{tan2018deep}, a very recent attention-based \cite{vaswani2017attention} BIO-tagging model that was developed concurrently with our work.
By adding the contextualized ELMo representations, we are able to out-perform all previous systems, including \newcite{peters2018deep}, which applies ELMo to the SRL model introduced in \newcite{He2017DeepSR}.

\begin{table*}[t]
\newcolumntype{Y}{>{\centering\arraybackslash}X}
\setlength{\tabcolsep}{.25em}
\newcommand\textrmlf[1]{{\NHLight#1}}
\centering
\footnotesize
\begin{tabularx}{\textwidth}{l c *{11}{Y}}
\toprule
& \multicolumn{4}{c}{CoNLL~05 In-domain (WSJ)} 
& \multicolumn{3}{c}{Out-of-domain (Brown)} 
& \multicolumn{4}{c}{CoNLL~2012 (OntoNotes)} \\
\cmidrule(lr){2-5} \cmidrule(lr){6-8} \cmidrule(lr){9-12} 
End-to-End & Dev. F1 & P  & R  & F1 & P & R & F1 & Dev. F1 &  P  & R  & F1 \\
\cmidrule{1-12}
\multicolumn{1}{l}{Ours+$\texttt{ELMo}$} & \textbf{85.3} & \textbf{84.8} & \textbf{87.2} & \textbf{86.0} & 
\textbf{73.9} & \textbf{78.4} & \textbf{76.1} &
\textbf{83.0} & \textbf{81.9} & \textbf{84.0} & \textbf{82.9}  \\
\multicolumn{1}{l}{\newcite{He2017DeepSR}$^{\texttt{PoE}}$}  & 81.5  & 82.0 & 83.4 & 82.7 & 69.7 & 70.5 & 70.1   &
77.2 & 80.2 & 76.6 & 78.4 \\
\cmidrule(lr){1-12}
\multicolumn{1}{l}{Ours}  & \textbf{81.6} & \textbf{81.2} & \textbf{83.9} & \textbf{82.5} &  \textbf{69.7} & \textbf{71.9} & \textbf{70.8}  & \textbf{79.4} &
\textbf{79.4}  & \textbf{80.1} & \textbf{79.8}  \\
\multicolumn{1}{l}{\newcite{He2017DeepSR}} &  80.3  & 80.2 & 82.3 & 81.2 & 67.6 & 69.6 & 68.5  &
75.5 & 78.6 & 75.1 & 76.8  \\
\bottomrule
\end{tabularx}
\caption{End-to-end SRL results for CoNLL~2005 and CoNLL~2012, compared to previous systems. CoNLL~05 contains two test sets: WSJ (in-domain) and Brown (out-of-domain).
}
\label{tab:srl_results_e2e}
\end{table*}

\begin{table}[t]
\newcolumntype{Y}{>{\centering\arraybackslash}X}
\newcommand\textrmlf[1]{{\NHLight#1}}
\centering
\footnotesize
\begin{tabularx}{\columnwidth}{l c *{3}{Y}}
\toprule
& \multicolumn{1}{c}{WSJ} 
& \multicolumn{1}{c}{Brown} 
& \multicolumn{1}{c}{OntoNotes} \\
\cmidrule{1-4}
\multicolumn{1}{l}{Ours+$\texttt{ELMo}$} &
\textbf{87.4} & \textbf{80.4} & \textbf{85.5} \\
\multicolumn{1}{l}{\newcite{peters2018deep}+$\texttt{ELMo}$} &
- & - & 84.6 \\
\multicolumn{1}{l}{\newcite{tan2018deep}$^{\texttt{PoE}}$} &
86.1 &  74.8 &  83.9 \\
\multicolumn{1}{l}{\newcite{He2017DeepSR}$^{\texttt{PoE}}$} &
84.6 & 73.6 & 83.4 \\
\multicolumn{1}{l}{\newcite{fitzgerald2015semantic}$^{\texttt{PoE}}$} &
80.3 & 72.2 & 80.1 \\
\cmidrule(lr){1-4}
\multicolumn{1}{l}{Ours} &
83.9 &  73.7 &  82.1 \\
\multicolumn{1}{l}{\newcite{tan2018deep}} &
\textbf{84.8} & \textbf{74.1} & \textbf{82.7} \\
\multicolumn{1}{l}{\newcite{He2017DeepSR}} &
83.1 &  72.1 &  81.7 \\
\multicolumn{1}{l}{\newcite{Yang2017AJS}} &
81.9 & 72.0 & - \\
\multicolumn{1}{l}{\newcite{zhou2015end}} &
82.8 & 69.4 & 81.1 \\
\bottomrule
\end{tabularx}
\caption{Experiment results with gold predicates.
}
\label{tab:srl_results_gold}
\end{table}

\section{Analysis}
Our model's architecture differs significantly from previous BIO systems in terms of both input and decision space. 
To better understand our model's strengths and weaknesses, we perform three analyses following \newcite{Lee2017EndtoendNC} and \newcite{He2017DeepSR}, studying (1) the effectiveness of beam pruning, (2) the ability to capture long-range dependencies, (3) agreement with syntactic spans, and (4) the ability to predict globally consistent SRL structures. The analyses are performed on the development sets without using ELMo embeddings.
\footnote{For comparability with prior work, analyses (2)-(4) are performed on the CoNLL~05 dev set with gold predicates.}

\begin{figure}[t]
\centering
\footnotesize
\begin{tabularx}{\linewidth}{lrlr}
\toprule
Arg. Beam  &  $\Phi_{\text{a}}$  & Pred. Beam  & $\Phi_{\text{p}}$      \\
\midrule
by ambulance& 2.5& says& 0.1\\
her mother ... ambulance& 2.2& transported& 0.0\\
her mother& 2.2& ambulance&-8.3\\
Priscilla& 1.9& been&-11.3\\
should& 1.8&&  \\
transported by ambulance&-0.3&&\\
Priscilla says .... ambulance&-2.2&&\\
ambulance&-3.2&&\\    
\midrule
\end{tabularx}
\includegraphics[width=0.9\columnwidth, keepaspectratio,trim={0.2cm 0cm 0.2cm 16.6cm},clip]{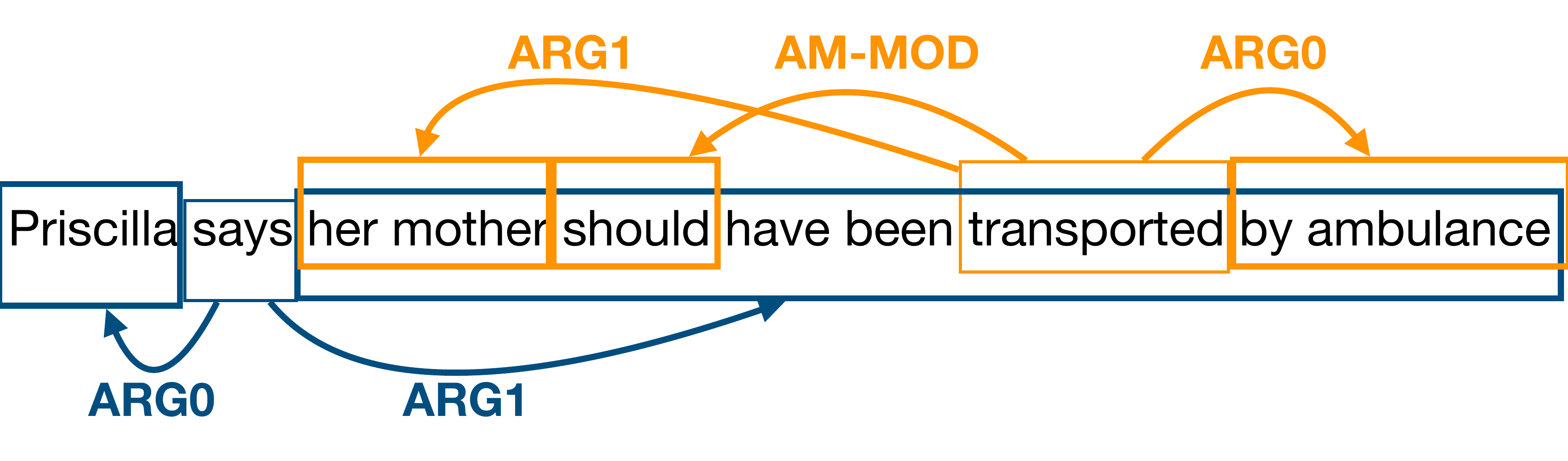}
\vspace{-1em}
\caption{Top: The candidate arguments and predicates in the argument beam $B_{\text{a}}$ and predicate beam $B_{\text{p}}$ after pruning, along with their unary scores. Bottom: Predicted SRL relations with two identified predicates and their arguments.}\label{fig:beams}
\end{figure}

\paragraph{Effectiveness of beam pruning}
Figure \ref{fig:beams} shows the predicate and argument spans kept in the beam, sorted with their unary scores.
Our model efficiently prunes unlikely argument spans and predicates, significantly reduces the number of edges it needs to consider. Figure \ref{fig:span_recall} shows the recall of predicate words on the CoNLL~2012 development set. By retaining $\lambda_{\text{p}} = 0.4$ predicates per word, we are able to keep over 99.7\% argument-bearing predicates. Compared to having a part-of-speech tagger (POS:X in Figure \ref{fig:span_recall}), our joint beam pruning allowing the model to have a soft trade-off between efficiency and recall.\footnote{The predicate ID accuracy of our model is not comparable with that reported in \newcite{He2017DeepSR}, since our model does not predict non-argument-bearing predicates.}

\paragraph{Long-distance dependencies}
Figure \ref{fig:longdist:f1} shows the performance breakdown by binned distance between arguments to the given predicates. Our model is better at accurately predicting arguments that are farther away from the predicates, even compared to an ensemble model~\cite{He2017DeepSR} that has a higher overall F1.
This is very likely due to architectural differences; in a BIO tagger, predicate information passes through many LSTM timesteps before reaching a long-distance argument, whereas our architecture enables direct connections between all predicates-arguments pairs.

\paragraph{Agreement with syntax}
As mentioned in \newcite{He2017DeepSR}, their BIO-based SRL system has good agreement with gold syntactic span boundaries (94.3\%) but falls short of previous syntax-based systems \cite{punyakanok2004semanticRL}. By directly modeling span information, our model achieves comparable syntactic agreement (95.0\%) to \newcite{punyakanok2004semanticRL} without explicitly modeling syntax.

\begin{figure}[t!]
\centering
\vspace{1em}
\begin{tikzpicture}
\begin{axis}[
    	width=0.95\columnwidth,
	    height=0.6\columnwidth,
	    legend cell align=left,
	    legend style={at={(1, 0)},anchor=south east,font=\scriptsize},
	    xtick={0.1, 0.2, 0.3, 0.4, 0.5, 0.6, 0.7},
   	 	ytick={10, 20, 30, 40, 50, 60, 70, 80, 90, 100},
   		ymin=51, ymax=103,
   		xtick pos=left,
   		xtick align=outside,
	    xmin=0.05,xmax=0.75,
	    mark options={mark size=3},
		font=\scriptsize,
   	 	ymajorgrids=true,
    	xmajorgrids=true,
    	xlabel=Spans per word $\lambda$,
        ylabel=Recall (\%),
    	ylabel style={yshift=-1ex,}]

\addplot[
    only marks,
    color=g-blue,
    mark=triangle*,
    mark size=15pt
    ]
    coordinates {
        (0.15, 88.95)
    };        
    \addlegendentry{POS:Verb}    

\addplot[
    only marks,
    color=g-blue,
    mark=square*,
    mark size=15pt
    ]
    coordinates {
    (0.42, 98.49)
    };        
    \addlegendentry{POS:Verb+Noun}    
    
\addplot[
    only marks,
    color=g-blue,
    mark=diamond*,
    mark size=15pt
    ]
    coordinates {
    (0.48, 99.67)
    };        
    \addlegendentry{POS:Verb+Noun+Adj}    
   	
\addplot+[
only marks,
    color=g-red,
    mark=o,
    line width=1.2pt,
    forget plot
    ]
    coordinates {
    (0.4, 99.7)
    };     
   
\addplot[
    color=g-red,
    line width=1.2pt
    ]
    coordinates {
    (0.1,51.71)(0.2,91.32)(0.3,98.4)(0.4, 99.71)(0.5, 99.95)(0.7, 100.0) 
    };
    \addlegendentry{Ours:Predicate beam}    
\end{axis}
\end{tikzpicture}
\caption{Recall of gold argument-bearing predicates on the CoNLL~2012 development data as we increase the number of predicates kept per word. POS:X shows the gold predicate recall from using certain pos-tags identified by the NLTK part-of-speech tagger \cite{bird2006nltk}. 
}
\label{fig:span_recall}
\vspace{1em}
\end{figure}
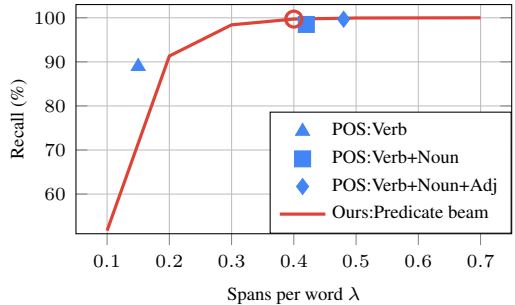
\begin{figure}[t]
\centering
\begin{tikzpicture}
\begin{axis}[
    	width=0.95\columnwidth,
	    height=0.55\columnwidth,
	    legend cell align=left,
	    legend style={at={(0, 0)},anchor=south west,font=\scriptsize},
	    mark options={mark size=2.5},
		font=\small,
	    xmin=-2,xmax=2.5,
   		ymin=56, ymax=88,
	    xticklabel style={align=center},
	    xtick=      {-1, 0, 1,  2  },
	    xticklabels={0,  1-2, 3-6, 7-max },
   	 	ytick={60, 65, 70, 75, 80, 85},
   	 	ymajorgrids=true,
    	xmajorgrids=true,
    	xlabel=Distance (num. words in between),
        ylabel=F1 \%,
    	ylabel style={yshift=-1ex,}]

    \addplot[smooth,mark=o,g-blue,line width=1.0pt] plot coordinates {
(-1,85.69	)
(0,	83.19	)
(1,	75.55	)
(2,	70.61	)
    };
    \addlegendentry{Ours}
    
    \addplot[smooth,mark=square,orange,line width=1.0pt] plot coordinates {
(-1,86.66	)
(0,	83.40	)
(1,	72.83	)
(2,	66.82	)
    };
    \addlegendentry{He (PoE)}
    
    \addplot[smooth,mark=triangle,g-red,line width=1.0pt] plot coordinates {
(-1,86.37	)
(0,	82.33	)
(1,	71.72	)
(2,	64.54	)
    };
    \addlegendentry{He}

    \addplot[smooth,mark=x,black,line width=1.0pt] plot coordinates {
(-1,	82.80	)
(0,	78.54	)
(1,	66.93	)
(2,	56.59	)
    };
    \addlegendentry{Punyakanok}
    
    \end{axis}
    
\end{tikzpicture}
\caption{
F1 by surface distance between predicates and arguments, showing degrading performance on long-range arguments.
\label{fig:longdist:f1}}
\end{figure}
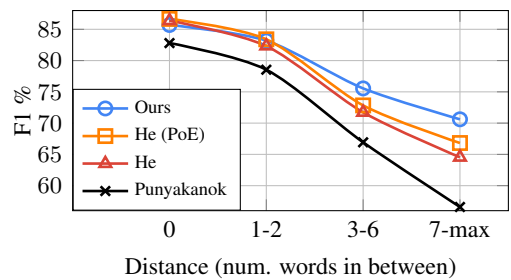

\paragraph{Global consistency}
On the other hand, our model suffers from global consistency issues.
For example, on the CoNLL~2005 test set, our model has lower complete-predicate accuracy (62.6\%) than the BIO systems \cite{He2017DeepSR,tan2018deep} (64.3\%-66.4\%).
Table \ref{tab:violations:srl} shows its violations of global structural constraints\footnote{\newcite{punyakanok2008importance} described a list of global constraints for SRL systems, e.g., there can be at most one core argument of each type for each predicate.} compared to previous systems. 
Our model made more constraint violations compared to previous systems. 
For example, our model predicts duplicate core arguments\footnote{Arguments with labels ARG0,ARG1,\ldots,ARG5 and AA.} (shown in the U column in Table \ref{tab:violations:srl}) more often than previous work. 
This is due to the fact that our model uses independent classifiers to label each predicate-argument pair, making it difficult for them to implicitly track the decisions made for several arguments with the same predicate.

The \emph{Ours+decode} row in Table \ref{tab:violations:srl} shows SRL performance after enforcing the U-constraint using dynamic programming \cite{tackstrom2015efficient} at decoding time. Constrained decoding at test time is effective at eliminating all the core-role inconsistencies (shown in the U-column), but did not bring significant gain on the end result (shown in SRL F1), which only evaluates the piece-wise predicate-argument structures.

\begin{table}[t]
\newcolumntype{Y}{>{\centering\arraybackslash}X}
\setlength{\tabcolsep}{.25em}
\centering
\small
\begin{tabularx}{\columnwidth}{l c *{5}{Y}}
\toprule
 & & & \multicolumn{3}{c}{SRL-Violations} \\ 
\cmidrule(lr){4-6}
  Model/Oracle   & SRL F1 & Syn \% & U & C & R  \\
\midrule
Gold    &  100.0  & 98.7    & 24     & 0     & 61          \\
\cmidrule{1-6}
Ours+decode  & 82.4 & 95.1  &  0   & 8  & 104 \\
Ours         & 82.3 & 95.0  &  69  & 7  & 105 \\
He (PoE)  & 82.7 & 94.3  &  37     & 3     & 68   \\
He        & 81.6 & 94.0  &  48     & 4     & 73   \\
Punyakanok  & 77.4 & 95.3 &   0  &   0   & 0  \\
\bottomrule
\end{tabularx}
\caption{Comparison on the CoNLL~05 development set against previous systems in terms of unlabeled agreement with gold constituency (Syn\%) and each type of SRL-constraints violations (\textbf{U}nique core roles, \textbf{C}ontinuation roles and \textbf{R}eference roles). 
} \label{tab:violations:srl}
\end{table}

\section{Conclusion and Future Work}
We proposed a new SRL model that is able to jointly predict all predicates and argument spans, generalized from a recent coreference system \cite{Lee2017EndtoendNC}. Compared to previous BIO systems, our new model supports joint predicate identification and is able to incorporate span-level features. Empirically, the model does better at long-range dependencies and agreement with syntactic boundaries, but is weaker at global consistency, due to our strong independence assumption. 

In the future, we could incorporate higher-order inference methods \cite{lee2018higher} to relax this assumption. It would also be interesting to combine our span-based architecture with the self-attention layers \cite{tan2018deep,strubell2018linguistically} for more effective contextualization.

\section*{Acknowledgments}

This research was supported in part by the ARO (W911NF-16-1-0121), the NSF  (IIS-1252835,  IIS-1562364), a gift from Tencent, and an Allen Distinguished Investigator Award. We thank Eunsol Choi, Dipanjan Das, Nicholas Fitzgerald, Ariel Holtzman, Julian Michael, Noah Smith, Swabha Swayamdipta, and our anonymous reviewers for helpful feedback. 


\bibliography{main}
\bibliographystyle{acl_natbib}

\appendix
\section{Supplemental Material}
\label{sec:supplemental}
\subsection{Hyperparameters}

\paragraph{Representation sizes}
The word embeddings are fixed 300-dimensional GloVe embeddings~\cite{Pennington2014GloveGV} (context window size of 2 for head word embeddings, and window size of 10 for LSTM inputs), normalized to be unit vectors. 
Out-of-vocabulary words are represented by a vector of zeros. In the character CNN, characters are represented as learned 8-dimensional embeddings. The convolutions have window sizes of 3, 4, and 5 characters, each consisting of 50 filters.

\paragraph{Network sizes}
We use 3 stacked bidirectional LSTMs with highway connections and 200 dimensional hidden states. Each MLP consists of two hidden layers with 150 dimensions and rectified linear units~\cite{relu}.

\paragraph{Inference}
We model spans up to length 30. We use $\lambda_{\text{a}} = 0.8$ for pruning arguments, $\lambda_{\text{p}} = 0.4$ for pruning predicates. At decoding time, we use dynamic programming (a simplified version of \newcite{tackstrom2015efficient}) to predict a set of non-overlapping arguments for each predicate \footnote{This is mainly a constraint enforced by the official CoNLL evaluation script.}.

\paragraph{Training}
We use Adam \cite{kingma:2016} with initial learning rate $0.001$ and decay rate of 0.1\% every 100 steps. The LSTM weights are initialized with random orthonormal matrices \cite{saxe:2013}. We apply 0.5 dropout to the word embeddings and character CNN outputs and 0.2 dropout to all hidden layers and feature embeddings. In the LSTMs, we use variational dropout masks that are shared across timesteps  \cite{gal:2016}, with $0.4$ dropout rate.

\paragraph{Batching}
At training time, we randomly shuffle all the documents and then batch at sentence level. Each batch contains at most $40$ sentences and $700$ words.
All models are trained for at most 320,000 steps with early stopping on the development set, which takes less than 48 hours on a single Titan X GPU.

\subsection{OntoNotes Data Statistics}\label{sec:data_split}

\begin{table}[t!]
\newcolumntype{Y}{>{\centering\arraybackslash}X}
\newcommand{\colindent}{\;}
\setlength{\tabcolsep}{0.25em}
\footnotesize
\centering
\begin{tabularx}{\linewidth}{l Y Y Y Y Y}
\toprule
 & \multicolumn{3}{c}{CoNLL~2012} & \multicolumn{2}{c}{OntoNotes5} \\
\cmidrule(lr){2-4}\cmidrule(lr){5-6}
 & Train & Dev & Test & Train & Dev  \\
\midrule
Docs & 2.8 & 0.3 & 0.3 & 11 & 1.5  \\
Sentences & 75 & 9.6 & 9.5 & 116 & 16  \\
Predicates & 189 & 24 & 24 & 253 & 35  \\
\bottomrule
\end{tabularx}
\caption{Data statistics (in number of thousands) for the CoNLL~2012 split and the train/dev split of OntoNotes5.
}
\label{tab:data_stats}
\vspace{-1em}
\end{table}

Table \ref{tab:data_stats} shows the data statistics on various splits of OntoNotes. 
We found that some sentences in the OntoNotes~5.0 train/dev split have missing predicates, which is unsuitable for training end-to-end SRL systems. Therefore, our end-to-end SRL models are trained on the smaller but cleaner CoNLL~2012 splits. For experiments with gold predicates, we use the full OntoNotes~5.0 train/dev split and the CoNLL~2012 test set, following previous work.

\end{document}